\documentclass[letterpaper, 10 pt, conference]{ieeeconf}

\IEEEoverridecommandlockouts                              
\makeatletter
\let\NAT@parse\undefined
\makeatother
\usepackage[numbers]{natbib}
\usepackage[utf8]{inputenc} 
\usepackage[T1]{fontenc}    
\usepackage{url}            
\usepackage{amsfonts}       
\usepackage{nicefrac}       
\usepackage{microtype}      
\usepackage{xcolor}         

\usepackage[bookmarks=true, colorlinks=true, citecolor=brown]{hyperref}
\usepackage{graphicx}        
\usepackage{booktabs}        
\usepackage{multirow}        
\usepackage{array}           
\usepackage{float}           
\usepackage{makecell}        
\usepackage{cuted}           
\usepackage{caption}         
\usepackage{wrapfig}
\usepackage{subcaption}
\usepackage{titletoc}

\usepackage{amsmath,amsfonts,bm}

\def\eqref#1{equation~\ref{#1}}

\def\1{\bm{1}}

\DeclareMathAlphabet{\mathsfit}{\encodingdefault}{\sfdefault}{m}{sl}
\SetMathAlphabet{\mathsfit}{bold}{\encodingdefault}{\sfdefault}{bx}{n}

\newcommand{\stdv}[1]{\scalebox{.95}{~$\pm$~#1}}

\title{\LARGE \bf
RetrDex: Efficient Object Retrieval in Cluttered Scenes \\ with a Dexterous Hand
}

\author{%
\authorblockN{%
\begin{tabular}[t]{@{}c@{}}
Fengshuo Bai$^{1,2,4}$, Yu Li$^{2,3}$, Jie Chu$^{1,2}$, Tawei Chou$^{2,3,4}$, Runchuan Zhu$^{5}$,\\
Ying Wen$^{1}$, Yaodong Yang$^{2,3}$, and Yuanpei Chen$^{2,3}$
\end{tabular}%
\thanks{Corresponding authors: Ying Wen ({\tt\small ying.wen@sjtu.edu.cn}) and Yaodong Yang ({\tt\small yaodong.yang@pku.edu.cn}).}}%
\authorblockA{%
\begin{tabular}[t]{@{}c@{}}
$^{1}$Shanghai Jiao Tong University,
$^{2}$PKU-PsiBot Joint Lab,
$^{3}$Peking University,\\
$^{4}$Zhongguancun Academy,
$^{5}$National University of Singapore
\end{tabular}}%
}

\begin{document}

\maketitle

\begin{strip}
    \centering
    \includegraphics[width=0.9\linewidth, height=0.37\linewidth]{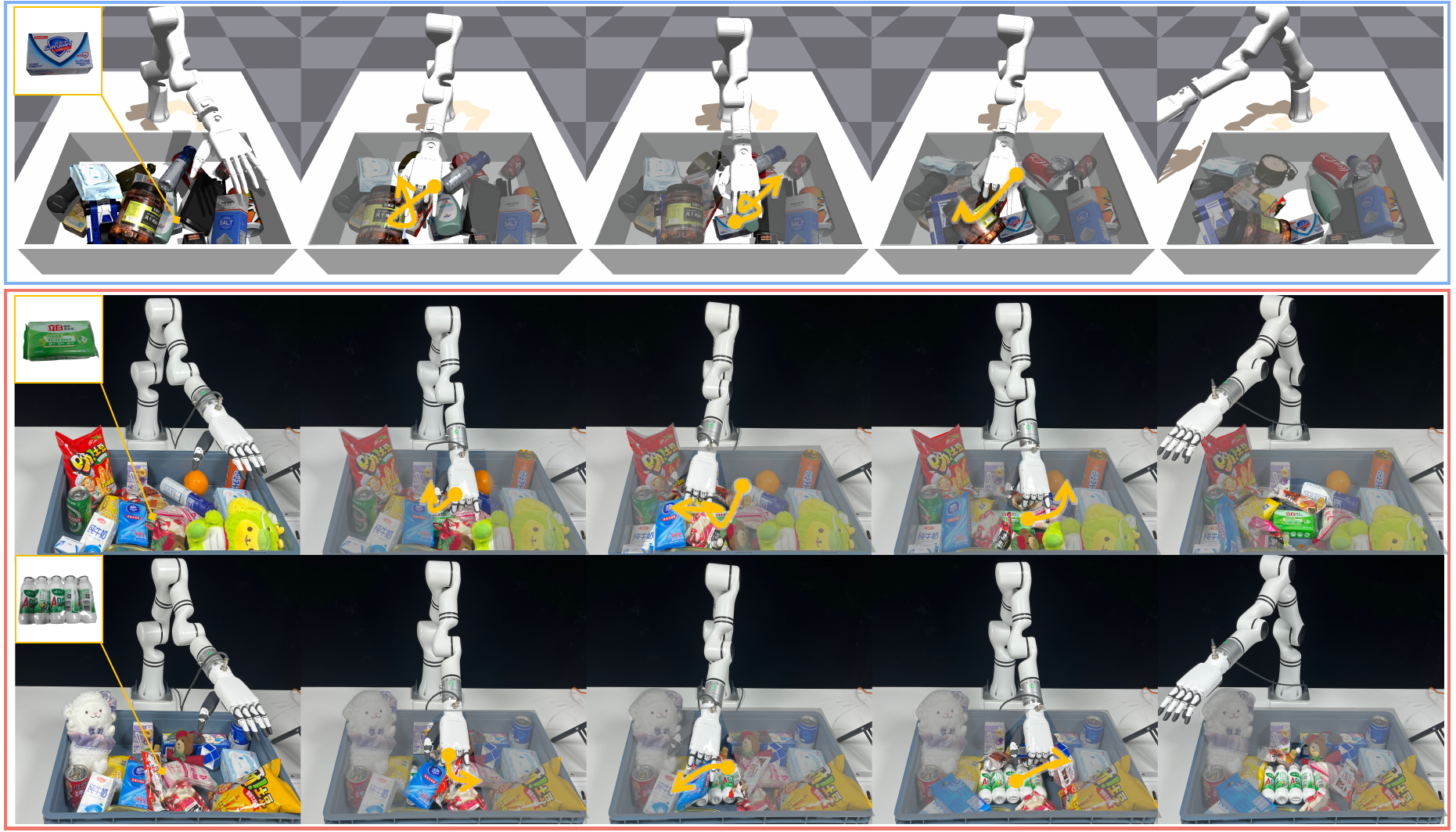}
    \captionof{figure}{\emph{RetrDex}, a framework that learns policies for efficient object retrieval in 
    \textbf{\textcolor[HTML]{82b0f8}{simulation}} and transfers zero-shot to \textbf{\textcolor[HTML]{f39890}{real-world}} deployment.}
    \label{fig:teaser}
    \vspace{-2ex}
\end{strip}

\begin{abstract}
Retrieving objects buried beneath clutter is both challenging and time-consuming, as complex support relationships make manipulation particularly difficult. Existing methods either focus on support relations and rely on sequential grasping to remove occluding objects, or perform preparatory actions such as pushing to facilitate subsequent grasps. However, these approaches are often inefficient and treat physical interactions as isolated auxiliary steps. In this paper, we propose \emph{RetrDex}, an efficient framework for dexterous arm-hand systems to learn object retrieval in cluttered scenes. Our approach leverages large-scale parallel reinforcement learning (RL) in diverse cluttered scenes and incorporates a spatially aware representation that encodes occlusion patterns and spatial relationships among the target, the dexterous hand, and surrounding clutter. This representation enables the policy to develop diverse manipulation skills (e.g., pushing, stirring, and poking) that actively clear occluders. We evaluate RetrDex on 16 household objects across varied clutter configurations, and obtain strong retrieval performance and efficiency on both seen and unseen targets. Furthermore, we demonstrate successful zero-shot transfer to a real-world dexterous multi-fingered robot system, validating the practical applicability of our method. Videos can be found on our project website:~\href{https://RetrDex.github.io}{https://RetrDex.github.io}.
\end{abstract}

\section{Introduction}
In everyday life, cluttered and unstructured environments are ubiquitous, and humans can skillfully search for and retrieve desired objects within them. Enabling robots to achieve similar capabilities is a key goal of embodied intelligence~\citep{JMLRv2219-804}. A major challenge arises when target objects are heavily occluded: in such cases, robots often cannot obtain crucial information, including precise poses or feasible grasp points, through vision or other non-contact sensing alone~\citep{8593986}. This limitation exposes a fundamental drawback of the classical perception-planning-action paradigm, which typically assumes access to a relatively complete world state before decision-making. To operate effectively in such environments, robots must move beyond passive observation and develop the ability to actively perceive and explore~\cite{8794143}.

To address these challenges, prior work has explored combining non-prehensile and prehensile actions, typically within hierarchical or sequential frameworks~\cite{11016017}. In such approaches, a policy first decides whether to perform preparatory actions such as pushing, which then create favorable conditions for subsequent grasping. While this paradigm improves success rates in cluttered scenes, it still treats physical interactions as isolated auxiliary steps before grasping. Owing to the limitations of parallel-jaw grippers, these pushes are often simplified into linear, goal-directed clearing motions. Even studies using dexterous hands~\cite{abs-2506-14317} mainly focus on learning a small set of discrete interaction-grasp combinations or on sequential removal of occluding objects~\cite{li2024broadcasting}. Despite these advances, current methods remain limited in modeling scene exploration and target retrieval as a continuous and deeply coupled process. In contrast, our work builds on a more fundamental assumption: the true advantage of dexterous hands lies in their capacity to learn a wide range of interactions, including stirring, poking, and reorienting, that extend beyond discrete pre-grasp motions. Our key insight is to enable the robot to autonomously discover efficient strategies for resolving occlusion, thereby overcoming the restrictions of sequential decision frameworks.

In this paper, we aim to develop an efficient retrieval policy for target objects in cluttered scenes. The policy is trained in simulation and achieves zero-shot sim-to-real transfer on a real robot. We construct a scene generation pipeline in IsaacGym~\citep{S2021_28dd2c79,bai2026safelab} that simulates natural clutter with strategically occluded targets. Building upon this, we introduce \emph{RetrDex}, a teacher-student framework for dexterous object retrieval. The teacher policy is trained in simulation with a spatially aware graph representation to capture spatial relationships among the hand, target, and surrounding objects. A graph neural network (GNN) learns this representation to encode interactions, enabling the teacher to reason about occlusions and learn effective retrieval strategies. The student policy is then distilled from the teacher via behavior cloning with a transformer architecture, which enhances generalization and supports zero-shot transfer to the real world.

We conduct extensive experiments in simulation and on a real robot. Results show our method retrieves objects successfully and improves efficiency over baselines. As shown in Figure~\ref{fig:teaser}, \emph{RetrDex} achieves zero-shot sim-to-real transfer.

\section{Related Work}
\textbf{Cluttered Object Manipulation.} Robotic manipulation in cluttered scenes has been extensively studied due to its importance for real-world applications such as household service and industrial automation~\cite{9197318}. Prior work has addressed this challenge from multiple perspectives: some studies improve robust grasping strategies~\cite{9197318,10342335}, others investigate retrieval tasks~\cite{li2024broadcasting}, while several works focus on rearrangement~\cite{pmlr-v205-tang23a} and dynamic actions such as grasping-and-throwing~\cite{kasaei2024harnessing}. Vision-based methods have also been widely explored; for instance, Huang et al.~\cite{9591286} predict object states after pushing to optimize grasping paths, and Kurenkov et al.~\cite{9341545} introduce a continuous pushing strategy guided by real-time visual feedback. Complementary deep RL methods further study multi-task policy learning~\cite{Bai_Zhang_Tao_Wu_Wang_Xu_2023} and preference-based optimization~\cite{bai2025star}. Despite these advances, most approaches target specific skills or restricted interaction modes, leaving open the challenge of developing unified policies that can flexibly address the diverse manipulation demands of cluttered scenes.

\textbf{Object Retrieval.} A key challenge within cluttered scenes is retrieving a specific target object, important in domestic service, logistics, and manufacturing. Planning-based methods focus on search and exploration, including optimization strategies~\cite{8793494} and teacher-guided exploration~\cite{9341545}, though they often rely on strong accessibility assumptions. Action-based approaches develop push-grasp synergies~\cite{9465702} or learn pushing and grasping without explicit foresight~\cite{8794143}, yet model interactions as discrete sequences. Structural reasoning methods analyze support relations for sequential removal of occluding objects~\cite{li2024broadcasting}. Perception-driven techniques incorporate tactile sensing~\cite{xu2024tactile} or multimodal visual-language reasoning~\cite{lemke2024spotcompose}. Hardware-focused efforts study different end-effectors, from non-prehensile pushing~\cite{8593986} and grippers~\cite{pmlr-v205-tang23a} to dexterous hands~\cite{pmlr-v229-chen23e}, with distinct trade-offs in control and dexterity. Our approach treats occlusion as the central difficulty and leverages a dexterous hand to actively manipulate clutter, modeling retrieval as a continuous and coupled process rather than an auxiliary step.

\section{Task Formulation}
We formulate object retrieval as a finite-horizon Markov decision process (MDP) defined by the tuple $(\mathcal{S}, \mathcal{A}, \mathcal{P}, \mathcal{R}, \gamma)$. Here, $\mathcal{S}$ and $\mathcal{A}$ denote the state and action spaces; $\mathcal{P}:\mathcal{S}\times\mathcal{A}\times\mathcal{S}\rightarrow[0,1]$ specifies the stochastic dynamics, i.e., the probability of transitioning to $s^\prime$ from state $s$ after action $a$; $\mathcal{R}:\mathcal{S}\times\mathcal{A}\times\mathcal{S}\rightarrow\mathbb{R}$ is the reward function; and $\gamma\in(0,1)$ is the discount factor. A policy $\pi(a|s)$ defines a distribution over actions given state $s$, aiming to maximize the expected return $\mathbb{E}_{\pi}\left[\sum_{t=0}^{T-1}\gamma^{t}r_t\right]$ over an episode of horizon $T$. In the context of cluttered retrieval, this formulation requires policies to acquire advanced skills for actively exploring clutter, removing occlusions, and exposing the targets, thereby enabling successful retrieval when the object is fully hidden.

\begin{figure*}
    \centering
    \includegraphics[width=0.9\linewidth,height=0.38\linewidth]{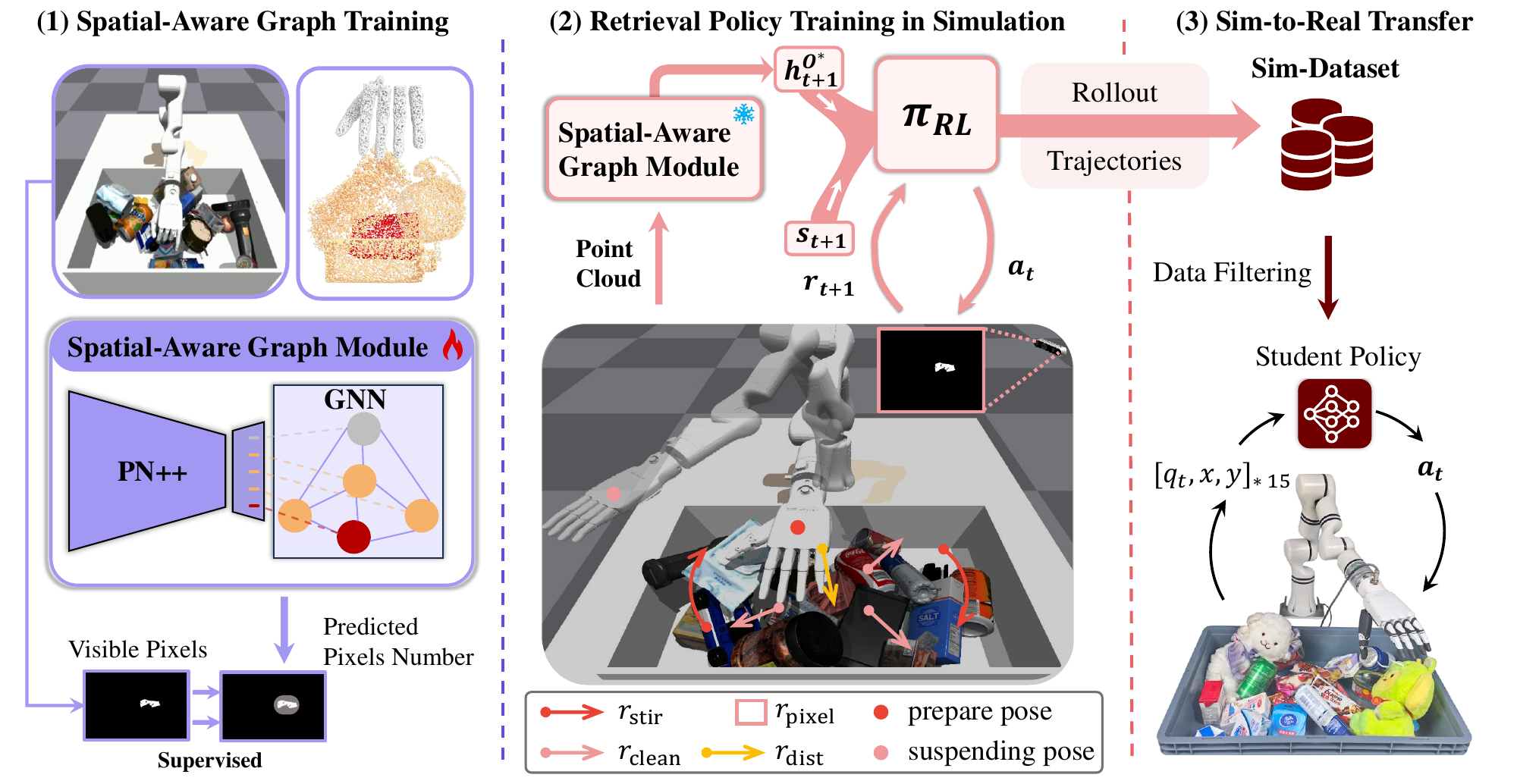}
    \vspace{-0.3em}
    \caption{\textbf{Overview of RetrDex.} 
    (a) A GNN module learns a spatially aware representation that encodes occlusion and spatial relationships among objects. 
    (b) Using this representation, a teacher policy is trained in simulation to acquire effective retrieval strategies. 
    (c) High-quality trajectories generated by the teacher are filtered and used to train a student policy via behavior cloning, enabling efficient deployment on a real robot. In (c), the student input $[(q_t^{arm}, x_t, y_t)]_{t-14:t}$ denotes a history of 15 timesteps of arm joint positions and target pixel coordinates in the top-down camera frame.}
    \label{fig:method}
    \vspace{-10pt}
\end{figure*}

\section{Method}
In this section, we present our system for efficient object retrieval, as illustrated in Figure~\ref{fig:method}. The framework comprises three main components: diverse clutter construction (Section~\ref{sec:task_con}), reinforcement learning (RL) problem formulation (Section~\ref{sec:rl_design}), and retrieval policy training (Section~\ref{sec:policy_train}). We describe sim-to-real policy transfer in Section~\ref{sec:sim2real}.

\subsection{Task Construction}\label{sec:task_con}
The main challenge in cluttered scenes arises from the diversity of object configurations, including categories, geometries, locations, poses and their combinations. To simulate realistic scenarios, we create cluttered scenes by dropping 20+ household objects of varying masses, sizes, and shapes into a box, with the target object placed at the bottom. For each episode in \emph{simulation}, both the target object and its pose are varied within the box boundaries.

To prevent hand interference during scene initialization and ensure reliable reward computation, we define two static robotic poses in the simulator: the \emph{prepare pose}, where the hand is positioned directly above the box and prepares to retrieve targets, and the \emph{suspending pose}, where the hand is kept away from the box to avoid affecting object dropping or reward evaluation. During the scene initialization stage, the hand moves to the \emph{suspending pose} while objects are released and then returns to the \emph{prepare pose} to begin training until objects remain static. During the training stage, the hand may occlude the target in the top-down camera view, so every 10 timesteps, we temporarily move it to the \emph{suspending pose} to compute the pixel reward item, defined as the number of target pixels visible in the segmentation mask, before restoring the previous configuration. This strategy stabilizes policy learning by evaluating cumulative behavior over time instead of relying on noisy immediate feedback. During policy evaluation, rewards are not computed, so this periodic hand movement is removed.

\begin{figure*}
    \centering
    \includegraphics[width=0.9\linewidth, height=0.25\linewidth]{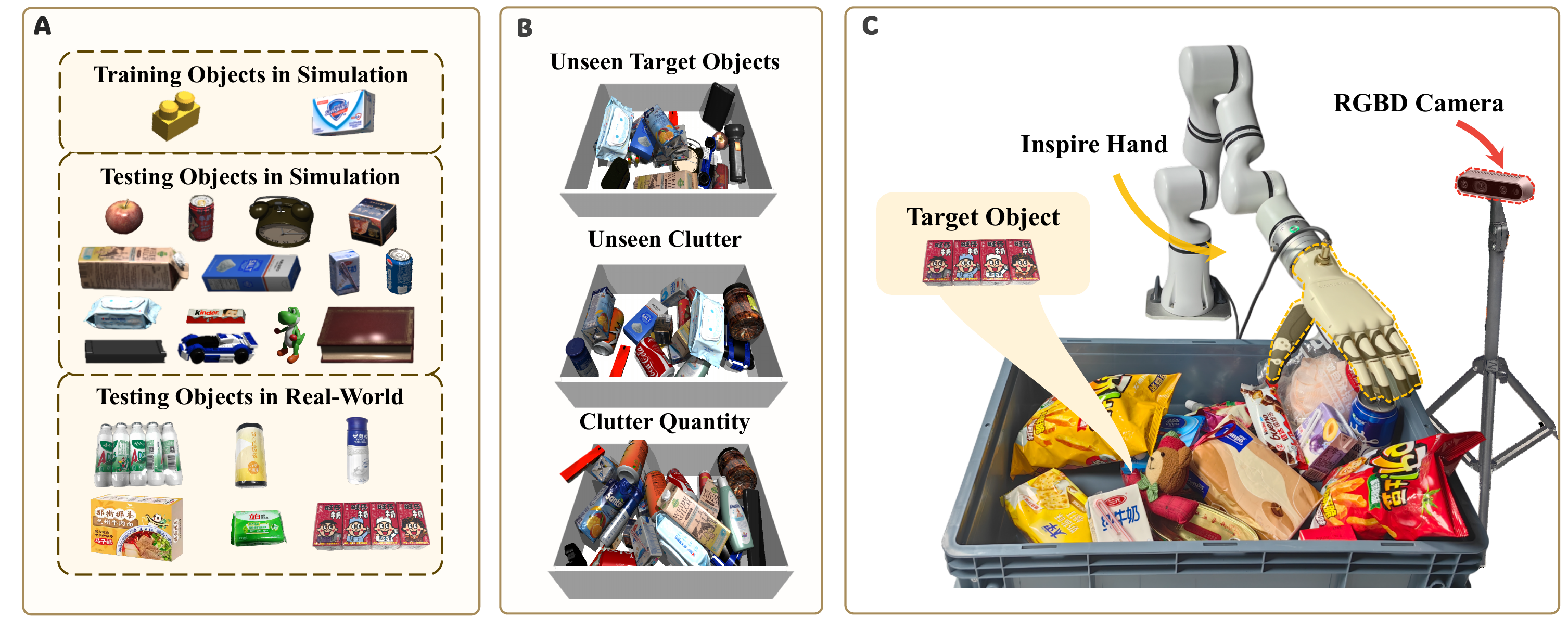}
    \vspace{-0.35em}
    \caption{\textbf{Experimental setups.}
    (a) Object sets for simulation training and for simulation/real-world testing.
    (b) Cluttered simulation scenes, including generalization settings L1, L2, and L3.
    (c) Real robot workspace with an Inspire Hand on a Realman RM-75 arm and a RealSense D435 camera.}
    \label{fig:dataset}
    \vspace{-10pt}
\end{figure*}

\subsection{RL Problem Design}\label{sec:rl_design}
After constructing the cluttered scenes, we address the object retrieval task using a model-free RL algorithm. We next describe the observation and action spaces of the policy, followed by the reward formulation.

\noindent \textbf{Observation Space.} At each timestep $t$, the control policy receives a combination of proprioceptive, visual, and privileged information. The proprioceptive inputs are joint positions $q_t = (q_t^{arm}, q_t^{hand}) \in \mathbb{R}^{13}$, where $q_t^{arm} \in \mathbb{R}^7$ denotes the arm joints and $q_t^{hand} \in \mathbb{R}^6$ the hand joints. The visual inputs are derived from the target object's segmentation mask, including its bounding box $b_t=(x_t,y_t,w_t,h_t) \in \mathbb{R}^4$, area $m_t=w_t \cdot h_t \in \mathbb{R}$, and the depth at the bounding-box center pixel $d_t \in \mathbb{R}$.

To facilitate policy learning in simulation, we additionally provide a spatially aware representation $h^{o^*}_t$ as privileged information, encoding occlusion and spatial relationships among the target, hand, and surrounding objects. The privileged information further consists of detailed kinematic states. Specifically, it includes the poses of the five fingertips $\{T_t^{f,i}\}_{i=1}^{5}$, where each $T_t^{f,i}\in \mathbb{R}^{7}$, and the pose of the target object $T_t^{obj}\in \mathbb{R}^{7}$. We also include the joint velocities $\dot{q}_t\in \mathbb{R}^{13}$, as well as the target object's linear and angular velocities $v_t^{obj}=(\mathbf{v}_t^{obj},\mathbf{\omega}_t^{obj})\in \mathbb{R}^{6}$. Finally, the positions of the five nearest objects to the target are represented as $\{T_t^{near,i}\}_{i=1}^{5}$, where each $T_t^{near,i}\in \mathbb{R}^{3}$. The complete observation is thus defined in Eq.~\eqref{eq:obs}:
\begin{equation}\label{eq:obs}
\small
s_t= \{q_t, b_t, m_t, d_t, \{T_t^{f, i}\}_{i=1}^{5}, T_t^{obj}, \dot{q}_t, v_t^{obj}, \{T_t^{near, i}\}_{i=1}^{5}, h^{o^*}_t\}.
\end{equation}

\noindent \textbf{Action Space.} The action space consists of the target joint positions of the robot, $a_t = (a_t^{arm}, a_t^{hand}) \in \mathbb{R}^{13}$. To improve control stability, the policy outputs raw target positions for the hand $\tilde{a}_t^{hand} \in \mathbb{R}^6$, which are smoothed with the previous target to reduce abrupt movements:  
\begin{equation}
a_t^{hand} = \lambda \tilde{a}_t^{hand} + (1-\lambda)a_{t-1}^{hand},
\end{equation}
where $\lambda$ is the smoothing factor. For the arm, the action $a_t^{arm}\in \mathbb{R}^7$ represents relative joint position changes, which are added to the current joint angles to compute the target positions used for control.

\noindent \textbf{Reward Function.} We design a reward function to optimize retrieval skills by encouraging the hand to actively expose the target object. The reward comprises five components. The \emph{distance reward} encourages the hand to approach occluded regions by minimizing the palm-to-target distance, defined as $r_\textrm{dist} = \exp(-5 \cdot \min(d - e_0, 0))$, where $d$ is the palm-to-target distance and $e_0=0.15$ is a threshold. The \emph{stir reward} encourages displacement of surrounding objects, especially under complete occlusion. Let $p^{all}_t$ denote the concatenated positions of all objects at timestep $t$; the reward is $r_\textrm{stir} = \alpha \|p^{all}_t - p^{all}_{t-1}\|_2$, with scaling factor $\alpha=400$. The \emph{proximity clearance reward} promotes clearing occluders near the target. Let $d_i$ be the distance between the target and its $i$-th nearest object, for $k$ neighbors; then $r_\textrm{clean} = \beta \cdot r_\textrm{dist} \cdot \sum_{i=1}^{k} d_i$, with weight $\beta=100$. The \emph{pixel visibility reward} provides a holistic vision-based signal by rewarding visible target pixels in the top-down segmentation mask. Let $C$ be the pixel count; then $r_\textrm{pixel} = C/15$. Finally, we add \emph{penalty} terms following prior work~\cite{lin2024twisting}, including action penalties, contact penalties, and penalties for displacing the target object.

To facilitate more efficient learning and reduce the risk of reward hacking~\cite{amodei2016concrete,Bai_Liu_Du_Wen_Yang_2025}, we adopt the potential-based reward shaping framework of Ng et al.~\cite{10.5555/645528.657613} to accelerate training. We define the potential function in Eq.~\eqref{eq:potential} and the total reward in Eq.~\eqref{eq:reward}:
\begin{equation}\label{eq:potential}
    \Phi(s) = r_\textrm{dist}(s) + r_\textrm{clean}(s) + r_\textrm{pixel}(s) + r_\textrm{stir}(s).
\end{equation}
\begin{equation}\label{eq:reward}
    \mathcal{R}(s, a, s^\prime) = \Phi(s^\prime) - \Phi(s) - r_\textrm{penalty}.\end{equation}

\subsection{Retrieval Policy Training}\label{sec:policy_train}
The training procedure consists of two stages. We first train a GNN module to predict target visibility, from which a spatially aware representation is extracted. Once trained, this module is frozen, and the learned representation is incorporated into the policy observation space to guide retrieval policy learning.

\noindent \textbf{Spatially Aware Representation Learning.}
To capture spatial interactions and occlusion relationships among the robotic hand, the target object, and surrounding clutter, we design a spatially aware representation module that integrates point cloud encoding with graph neural networks. We construct a graph $\mathcal{G}=(\mathcal{V},\mathcal{E})$, where $\mathcal{V}$ includes the target object, the robotic hand, and its $K$ nearest neighboring objects, yielding $K+2$ nodes. Each node is represented by a sampled point cloud $\mathcal{P}_i\in\mathbb{R}^{256\times 3}$, encoded into a feature vector $e_i\in\mathbb{R}^{64}$ using PointNet++~\cite{DBLP:conf/nips/QiYSG17}. The node embeddings are connected in a fully connected, undirected graph, and two layers of EdgeConv~\cite{WangSLSBS19} are applied to propagate features and model context. The embedding of the target node after message passing, denoted $h^{o^*}_t$, serves as a spatially aware representation encoding occlusion and structural context.

The module is trained by regressing predicted visibility $\hat{v}$ against ground-truth $v$ using mean squared error, where visibility is the number of target pixels visible in the segmentation mask from the top-view camera. The learned representation $h^{o^*}_t$ is then injected into the policy observation space to enable spatial reasoning during manipulation.

\noindent \textbf{Retrieval Policy.} 
We train a closed-loop retrieval policy using Proximal Policy Optimization (PPO)~\cite{schulman2017proximal} to control a dexterous hand for retrieving occluded targets. Leveraging the massively parallel simulation capabilities of IsaacGym~\cite{S2021_28dd2c79}, we train across 512 environments in parallel. To improve robustness and generalization, we apply domain randomization by varying target objects, target poses, target mass, clutter layouts, and camera positions. At each episode start, all objects, including the target, are randomly initialized by dropping them into a box, generating diverse and challenging clutter configurations. Additionally, we randomize the initial pose and mass of the target object to enhance policy robustness. 

{
\setlength{\extrarowheight}{1.0pt}
\begin{table*}[!t]
\centering
\caption{
\small
\textbf{Main results across all methods.} We report retrieval success rate (RSR), retrieval steps (RS), and Increase in Exposure Ratio (IER) on three settings. All results are averaged over 5 seeds, each with 64 evaluation trajectories. Our method outperforms all baselines in both success rate and efficiency across all settings, demonstrating strong generalization and policy effectiveness.
}
\vspace{-0.6em}
\label{tab:main_res}
\renewcommand{\arraystretch}{1.0}
\resizebox{0.99\linewidth}{!}{%
\begin{tabular}{cccccccccccc}
\toprule
\multirow{2}{*}{Method} & \multicolumn{3}{c}{Seen Objects} && \multicolumn{3}{c}{Unseen Small Objects} && \multicolumn{3}{c}{Unseen Large Objects} \\ \cline{2-4} \cline{6-8} \cline{10-12} 
                                 & RSR ($\uparrow$)  &  RS ($\downarrow$)  & IER ($\uparrow$) &&  RSR ($\uparrow$)  &  RS ($\downarrow$)  & IER ($\uparrow$) &&  RSR ($\uparrow$)  &  RS ($\downarrow$)  & IER ($\uparrow$) \\ \midrule
VMP                              & 25.3\stdv{5.8}& 192.3\stdv{1.7} & 61.2\stdv{1.3}&& 32.2\stdv{3.9}& 179.3\stdv{6.3} & 60.4\stdv{2.6}&& 8.3 \stdv{3.2}& 204.5\stdv{2.1}  & 51.0\stdv{0.9}   \\
Grasp-Pick                       & 48.9\stdv{9.8}& 151.2\stdv{10.2}& 72.4\stdv{4.3}&& 24.3\stdv{9.1}& 181.4\stdv{7.1} & 54.9\stdv{5.6}&& 3.2 \stdv{2.1}& 207.2\stdv{7.1}  & 26.7\stdv{3.5}   \\
Ours (SAC)                  & 80.2\stdv{7.0}& 107.2\stdv{9.4} & 76.7\stdv{1.3}&& 68.8\stdv{5.6}& 135.0\stdv{2.1} & 75.5\stdv{3.7}&& 9.9 \stdv{3.7}& 201.4\stdv{4.6}  & 52.0\stdv{0.5}   \\
Ours (Gripper)              & 52.4\stdv{5.7}& 122.1\stdv{3.0} & 76.9\stdv{2.3}&& 45.0\stdv{5.8}& 142.9\stdv{7.5} & 61.1\stdv{1.3}&& 19.2\stdv{1.4}& 174.5\stdv{3.9}  & 62.7\stdv{3.0}   \\ \cline{1-12}
\textbf{Ours}               & 91.4\stdv{3.1}& 91.2 \stdv{2.1} & 96.2\stdv{1.2}&& 86.2\stdv{2.1}& 107.8\stdv{3.5} & 89.8\stdv{4.1}&& 73.9\stdv{3.1}& 133.2\stdv{4.2}  & 89.0\stdv{1.8}   \\ \bottomrule
\end{tabular}%
}
\vspace{-10pt}
\end{table*}

}

\subsection{Sim-to-Real Transfer}\label{sec:sim2real}
When deploying policies in the real world, the teacher policy is not executed directly. Instead, we collect rollout trajectories from the trained teacher and distill a student policy for deployment. We select successful trajectories where the fingers avoid contact with the box bottom (maintaining at least 2 cm clearance) and target objects are evenly distributed, ensuring stable demonstrations. The student policy is trained via behavior cloning (BC)~\cite{NIPS1988_812b4ba2}. Its observation space is $o_t = (q_t^{arm}, x_t, y_t)$, consisting of arm joint positions and the target object's pixel coordinates in the top-down camera frame. Joint positions are directly accessible, whereas the target object is tracked using a top-down camera: the Segment Anything Model (SAM)~\cite{kirillov2023segment} provides initial segmentation, and Cutie~\cite{cheng2024putting} ensures continuous tracking. The object's pixel coordinates are converted into world coordinates via camera calibration at 30 Hz. The student policy outputs actions $a_t \in \mathbb{R}^{13}$, corresponding to the target joint positions of both the arm and the dexterous hand. As illustrated in Figure~\ref{fig:method}, we implement the policy as a transformer network that processes a history of 15 timesteps of $(q_t^{arm}, x_t, y_t)$, enabling it to capture temporal dependencies essential for retrieval in cluttered environments.

\section{Experiments}
We conduct experiments in simulation and real-world settings. We first examine overall performance and the dexterous hand comparison (Sec.~\ref{subsec:q12}), then ablations (Sec.~\ref{subsec:q34}), generalization and efficiency (Sec.~\ref{subsec:q567}), and real-robot transfer (Sec.~\ref{subsec:q8}). All simulation results are averaged over 5 random seeds with 64 trajectories per seed, and real-world results are collected from 10 independent trials per setting. Simulation experiments use the privileged teacher policy (Sec.~\ref{sec:policy_train}), while real-world deployment uses the distilled student policy (Sec.~\ref{sec:sim2real}).

\subsection{Experimental Setup}
\noindent \textbf{Dataset.} During retrieval policy optimization, we use a LEGO block and a soap box as target objects in the training dataset. At the beginning of each episode, one of these objects is randomly selected for policy training. To evaluate the generalizability of our policy to unseen objects, we supplement a test set comprising 8 small objects (e.g., apples) and 6 large objects (e.g., books), differentiated by shape and weight. Furthermore, we introduce several new objects to construct novel clutter, testing the policy’s ability to generalize across various cluttered environments.

\noindent \textbf{Evaluation Metrics.} We evaluate the retrieval policy in terms of both effectiveness and efficiency, based on the pixel-level visibility of the target object. The core metric, \emph{exposure}, is defined as the ratio between the number of visible pixels of the target object under occlusion ($p_t^{curr}$) and its fully visible counterpart ($p_t^{all}$) under the same 6D pose. Specifically, we record the target object’s pose and visible pixel count every 10 steps during the retrieval episode. After each episode, all occluding objects are removed in simulation, and the target is rendered at the recorded pose to compute $p_t^{all}$. The exposure at time $t$ is computed as ${exposure}_t = p_t^{curr} / p_t^{all}$. 

A retrieval is deemed successful if the final exposure exceeds 95\% within 210 steps. We report three metrics: \emph{Retrieval Success Rate (RSR)}, defined as the percentage of successful trials; \emph{Retrieval Steps (RS)}, the number of steps to reach the 95\% exposure threshold; and \emph{Increase in Exposure Ratio (IER)}, which measures the absolute improvement in exposure from the initial to the final timestep. 

\noindent \textbf{Baseline Methods.} We evaluate our method against several competitive baselines designed to isolate the effects of algorithm choice, end-effector type, and planning strategy. \emph{Ours} uses PPO and a dexterous hand trained with a task-specific reward function. \emph{Ours (SAC)} replaces PPO with Soft Actor-Critic (SAC)~\cite{haarnoja2018soft}, keeping all other components identical. \emph{Ours (Gripper)} substitutes the dexterous hand with a parallel-jaw gripper to assess the impact of reduced manipulation flexibility. \emph{Visual-based Motion Planning (VMP)} is a heuristic approach that uses segmentation masks to guide a scripted motion planner with hand-designed rules. \emph{Grasp-Pick} assumes known stacking relationships of occluding objects and sequentially grasps and removes them using ground-truth object poses until the target is exposed.

{
\setlength{\extrarowheight}{1.pt}
\begin{table*}[!t]
\centering
\caption{
\small
\textbf{Ablation study on reward design and GNN module.} We report retrieval success rate (RSR), retrieval steps (RS), and Increase in Exposure Ratio (IER) on three settings. All results are averaged over 5 seeds, each with 64 evaluation trajectories.
}
\vspace{-0.6em}
\label{tab:ablation_res}
\resizebox{0.99\linewidth}{!}{%
\begin{tabular}{lccccccccccc}
\toprule
\multirow{2}{*}{Method}    & \multicolumn{3}{c}{Seen Objects}                        && \multicolumn{3}{c}{Unseen Small Objects}                && \multicolumn{3}{c}{Unseen Large Objects}                 \\ \cline{2-4} \cline{6-8} \cline{10-12} 
                           & RSR ($\uparrow$) & RS ($\downarrow$) & IER ($\uparrow$) && RSR ($\uparrow$) & RS ($\downarrow$) & IER ($\uparrow$) &&  RSR ($\uparrow$) & RS ($\downarrow$) & IER ($\uparrow$) \\ \midrule
\textbf{Ours(full)}        & 91.4\stdv{3.1}   & 91.2 \stdv{2.1}   & 96.2\stdv{1.2}   && 86.2\stdv{2.1}   & 107.8\stdv{3.5}   & 89.8\stdv{4.1}   && 73.9\stdv{3.1}    & 133.2\stdv{4.2}   & 89.0\stdv{1.8}   \\
\quad w/o Shaping           & 55.4\stdv{1.6}   & 149.4\stdv{2.8}   & 73.8\stdv{2.9}   && 46.7\stdv{2.6}   & 158.7\stdv{10.8}  & 69.3\stdv{3.7}   && 26.2\stdv{4.8}    & 179.5\stdv{1.7}   & 67.4\stdv{1.0}   \\
\quad w/o $r_\text{stir}$  & 73.8\stdv{1.4}   & 132.4\stdv{9.4}   & 84.6\stdv{1.3}   && 69.2\stdv{6.0}   & 140.2\stdv{4.8}   & 79.7\stdv{3.6}   && 34.9\stdv{7.2}    & 174.3\stdv{5.6}   & 76.5\stdv{1.0}   \\
\quad w/o $r_\text{clean}$ & 69.2\stdv{3.3}   & 126.2\stdv{2.2}   & 79.4\stdv{1.2}   && 63.5\stdv{3.8}   & 134.7\stdv{4.6}   & 72.8\stdv{1.3}   && 39.0\stdv{5.8}    & 167.9\stdv{1.4}   & 73.9\stdv{1.7}   \\ \midrule 
\quad w/o GNN              & 76.4\stdv{3.5}   & 117.3\stdv{4.6}   & 85.2\stdv{1.9}   && 68.3\stdv{2.2}   & 143.1\stdv{6.7}   & 77.4\stdv{2.1}   && 49.5\stdv{3.1}    & 169.2\stdv{5.0}   & 77.3\stdv{1.8}   \\
\quad w GNN (3-node)       & 81.2\stdv{2.7}   & 110.1\stdv{4.0}   & 88.0\stdv{1.6}   && 83.4\stdv{2.5}   & 112.2\stdv{5.1}   & 87.3\stdv{2.0}   && 51.1\stdv{2.9}    & 164.9\stdv{4.7}   & 81.5\stdv{1.6}   \\
\quad w GNN (7-node)       & 85.0\stdv{2.9}   & 104.3\stdv{3.5}   & 90.3\stdv{1.7}   && 78.2\stdv{2.0}   & 126.5\stdv{6.2}   & 85.2\stdv{1.8}   && 63.5\stdv{3.2}    & 155.7\stdv{6.3}   & 85.1\stdv{1.4}   \\ \bottomrule
\end{tabular}%
}
\vspace{-10pt}
\end{table*}
}

\subsection{Overall Performance}\label{subsec:q12}
\noindent \textbf{Main Results.}
We evaluate all methods on both seen and unseen targets, where the unseen category is further divided into small and large objects. As shown in Table~\ref{tab:main_res}, our method consistently achieves higher retrieval success rates (RSR) and requires fewer retrieval steps (RS) across all scenarios. Relative to the strongest baseline in each category, our method improves unseen RSR by 25.3\% on small targets (vs.\ SAC) and 285.0\% on large targets (vs.\ the gripper variant), indicating good generalization across object scales.

In contrast, VMP relies on segmentation masks and predefined open-loop rules, which limits adaptation to diverse clutter and yields lower success rates and efficiency. Grasp-Pick decomposes retrieval into a sequence of grasp-and-remove operations. Even with oracle knowledge of object stacking, error accumulation and long-horizon inefficiency reduce its performance. Replacing PPO with SAC in our framework, the trained policy occasionally exhibits emergent behavior such as directly grasping and lifting the target object. However, despite its promising training performance, it generalizes poorly to unseen target objects. We hypothesize that SAC's maximum-entropy objective encourages broader exploration, which may overfit to training objects and reduce robustness to novel targets. Overall, our policy achieves high success rates on both seen and unseen target objects.

\noindent \textbf{Performance on Dexterous Hand.} 
We also investigate whether dexterous hands offer an advantage over parallel-jaw grippers in cluttered retrieval tasks. We compare our full system (\emph{Ours}) with a variant using a gripper (\emph{Ours (Gripper)}), keeping the policy structure, reward design, and training settings identical. As shown in Table~\ref{tab:main_res}, \emph{Ours} achieves a success rate of 91.4\% on seen objects and 80.0\% on unseen objects on average over unseen small and large targets, compared with 52.4\% and 32.1\% for the gripper baseline. It also completes tasks with fewer steps. The gripper variant often pushes the target deeper or becomes stuck on surrounding items due to its limited motion capability. The dexterous hand can perform in-place poking, stirring, and fine-scale lifting, which leads to better clearance strategies.

\subsection{Ablation Studies}\label{subsec:q34}
\noindent \textbf{Effectiveness of the GNN Module.} To evaluate the contribution of our spatially aware graph module, we conduct ablation studies comparing policies with and without the GNN representation. Results in Table~\ref{tab:ablation_res} indicate that incorporating the GNN module consistently improves retrieval success. We further analyze the effect of varying the number of nearest neighbor objects $K$ included in the graph. As shown in Table~\ref{tab:ablation_res}, using $K=3$ yields good performance in retrieving small target objects, striking a balance between context richness and representation sparsity. For larger targets, the surrounding context becomes more complex; in these cases, the marginal gains from the GNN diminish as local interactions alone may not fully capture the occlusion relationships. Increasing $K$ to 7 introduces more spatial relationships with higher computational overhead and potential over-smoothing due to densely connected graphs. This leads to degraded performance, likely caused by the inclusion of redundant obstacles that dilute the target-centric representation.

\noindent \textbf{Impact of Reward Design.} To further assess the impact of reward design, we conduct an ablation study on individual reward components. As shown in Table~\ref{tab:ablation_res}, removing the stir reward ($r_{stir}$), which drives the displacement of surrounding clutter, leads to a 52.8\% drop in success rate on large objects, indicating its importance for resolving deep occlusions. In contrast, ablating the proximity clearance reward ($r_{clean}$), responsible for local decluttering near the target, causes a notable decline in performance on smaller objects, where tight spatial arrangements are common. Moreover, excluding the potential-based shaping function destabilizes learning, increasing retrieval steps and reducing overall success, whereas its inclusion improves success rate by 65.0\% and reduces steps by 39.0\% on seen objects. These results confirm the necessity of well-structured rewards for learning effective object retrieval strategies under occlusion.

\begin{figure*}[!t]
    \centering
    \subfloat[Generalization Summary\label{fig:gen_23}]{
        \includegraphics[width=0.26\textwidth]{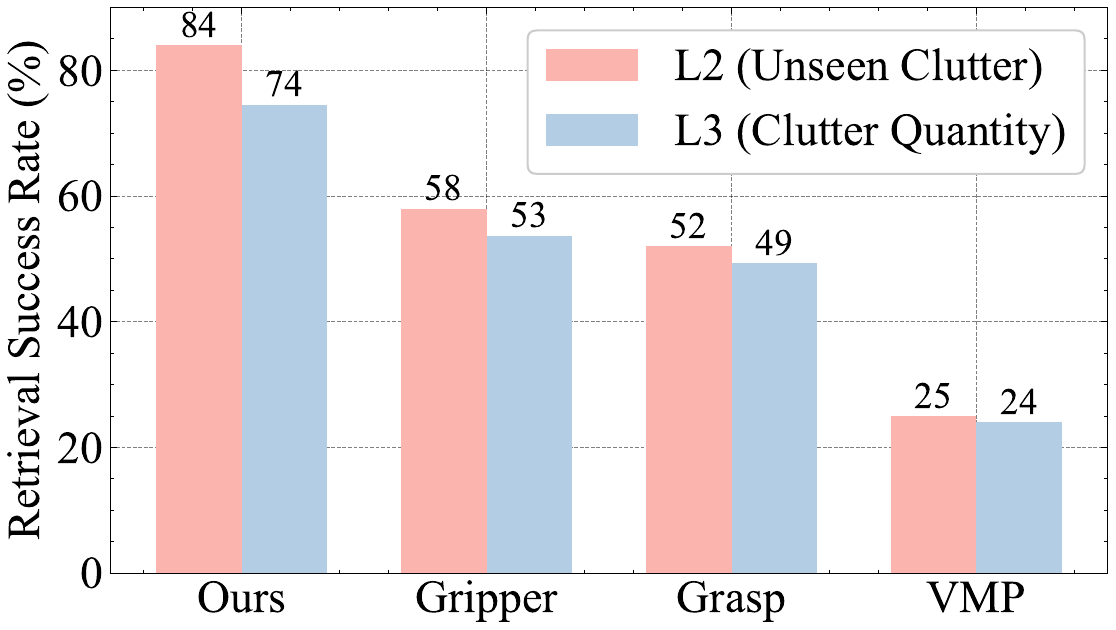}
    }
    \subfloat[Simulation Efficiency \label{fig:retrieval_steps}]{
        \includegraphics[width=0.26\textwidth]{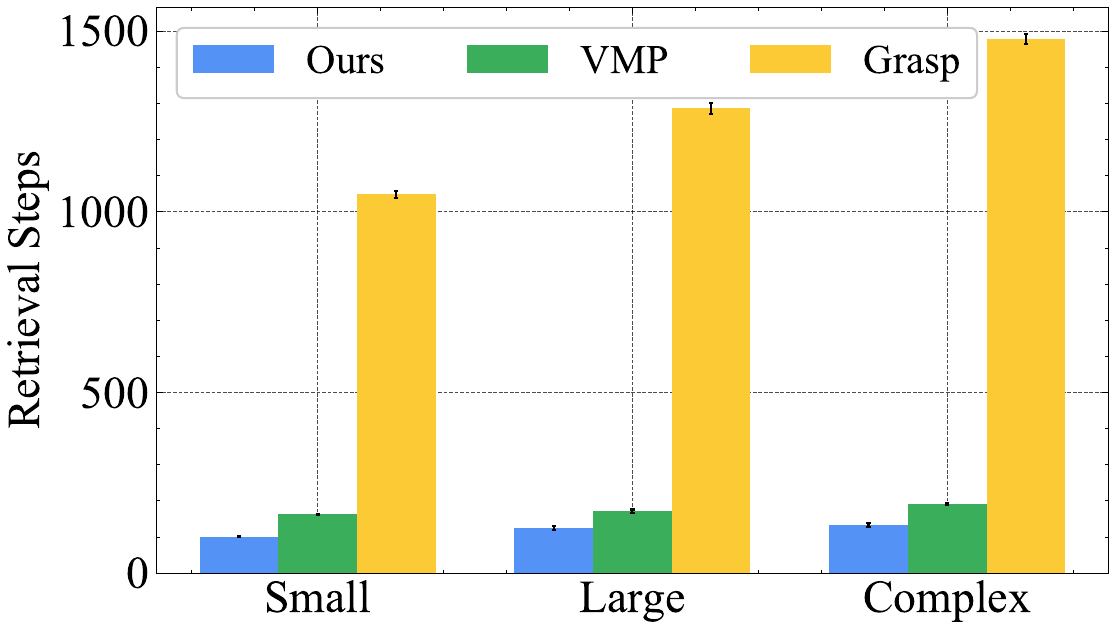}
    }
    \subfloat[Real-World Efficiency \label{fig:real_efficiency}]{
        \includegraphics[width=0.26\textwidth]{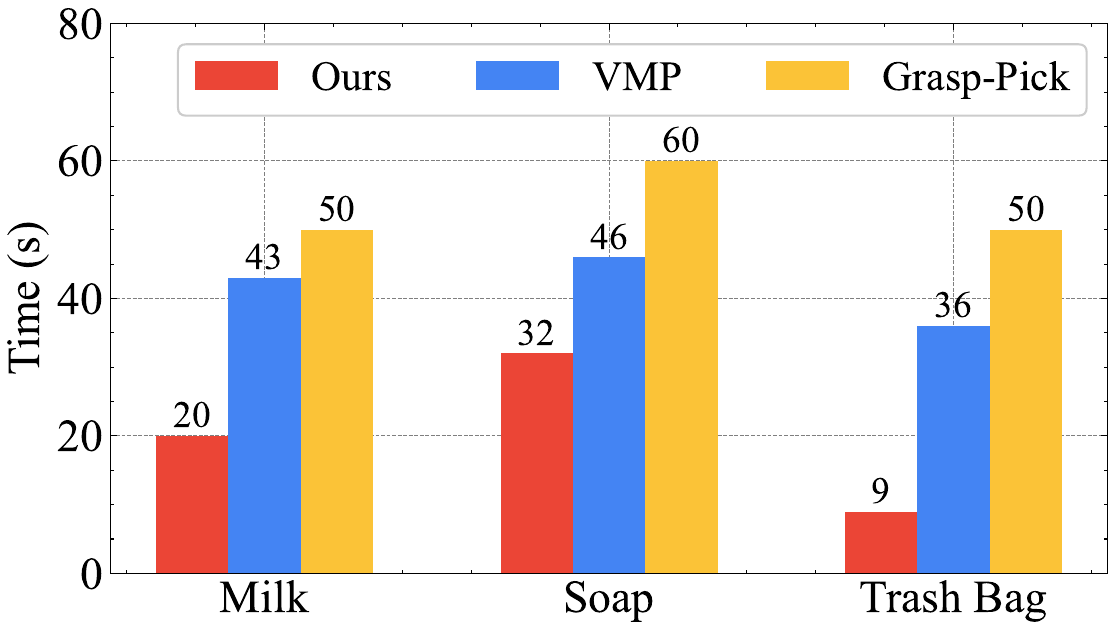}
    }
    \vspace{-0.5em}
    \caption{
    \textbf{Generalization and efficiency results.}
    (a) Simulation: success rates under unseen clutter (L2) and increased clutter (L3).
    (b) Simulation: retrieval steps for successful trials.
    (c) Real world: completion time for successful trials.
    }
    \label{fig:novel_behavior}
\vspace{-10pt}
\end{figure*}


\begin{table*}[t]
  \centering
  \begin{subtable}[t]{0.49\textwidth}
    \centering
    \caption{Performance on various target objects.}
    \label{tab:real_target_generalization}
    \renewcommand{\arraystretch}{1.}
    \resizebox{\linewidth}{!}{
    \begin{tabular}{clllllll}
    \toprule
    \multirow{2}{*}{Setting} & \multicolumn{4}{c}{Cuboid} & \multicolumn{2}{c}{Cylinder} & Sphere \\
    \cmidrule(lr){2-5} \cmidrule(lr){6-7} \cmidrule(lr){8-8}
    & Milk1 & Milk2 & Noodles & Soap & Yogurt & Bag & Ball \\
    \midrule
    \textbf{Ours} & 6/10 & 8/10 & 7/10 & 9/10 & 7/10 & 9/10 & 8/10 \\
    VMP          & 2/10 & 3/10 & 0/10 & 1/10 & 1/10 & 4/10 & 4/10 \\
    \bottomrule
    \end{tabular}
    }
  \end{subtable}
  \hfill
  \begin{subtable}[t]{0.49\textwidth}
    \centering
    \caption{Success rates across object positions.}
    \label{tab:real_position_generalization}
    \renewcommand{\arraystretch}{1.}
    \resizebox{\linewidth}{!}{
    \begin{tabular}{cccccc}
    \toprule
    Position & Center & Top Left & Bottom Left & Top Right & Bottom Right \\
    \midrule
    \textbf{Ours} & 8/10 & 6/10 & 7/10 & 5/10 & 6/10 \\
    VMP          & 2/10 & 1/10 & 3/10 & 0/10 & 2/10 \\
    \bottomrule
    \end{tabular}
    }
  \end{subtable}
  \caption{Real-robot performance comparison under different object types (left) and positions (right).}
\vspace{-17pt}
\end{table*}
\vspace{-0.6em}

\subsection{Additional Evaluations}\label{subsec:q567}
\noindent \textbf{Generalization to Unseen Clutter and Clutter Quantity.}
We evaluate the generalization capability of all methods under three settings of increasing difficulty, as illustrated in Figure~\ref{fig:dataset}(b). These include unseen target object generalization (L1), unseen clutter configuration generalization (L2), and increased clutter quantity generalization (L3). In L1, the target object is replaced while the clutter remains fixed. In L2, 80\% of the clutter objects are changed while keeping the target object constant. In L3, 50\% more clutter items are added to increase scene complexity. As shown in Table~\ref{tab:main_res}, our method demonstrates strong generalization across target object shapes. Furthermore, Figure~\ref{fig:gen_23} shows that our policy maintains high success rates even in L2 and L3 scenarios. In contrast, the performance of Grasp-Pick drops sharply as clutter density increases, due to its reliance on accurate sequential removals under occlusion. Similarly, gripper-based policies struggle to generalize across diverse clutter layouts, leading to frequent failures in both retrieval efficiency and success rates.

\noindent \textbf{Retrieval Efficiency.}
We evaluate the retrieval efficiency of our system by comparing the number of steps required to successfully retrieve the target object against several baselines. Experiments are conducted under three conditions: small targets, large targets, and complex clutter, defined as environments with 50\% more clutter than standard scenes. All settings involve target objects with 90\% occlusion, a level chosen to evaluate performance under challenging clutter. Our method leverages a dexterous multi-finger hand capable of skill-based manipulations such as pushing, stirring, and poking, which allows it to directly interact with occluders instead of sequentially removing them. As shown in Figure~\ref{fig:retrieval_steps}, our approach achieves higher efficiency than the baselines, reducing the number of steps by an average of 38\% compared to VMP and by 90\% compared to Grasp-Pick.

\subsection{Real-World Experiments}\label{subsec:q8}
We conduct sim-to-real experiments with the student policy on a real robot, as shown in Figure~\ref{fig:dataset}(c). Targets are tracked online with SAM and Cutie (Sec.~\ref{sec:sim2real}). We compare against VMP and sequential grasp-and-remove baselines.

\noindent \textbf{Retrieving Diverse Target Objects.}
We evaluate the retrieval performance of our policy on everyday objects with different shapes and sizes. As summarized in Table~\ref{tab:real_target_generalization}, we consider cuboids, cylinders, and spheres, covering both small and large instances. Our method achieves high success rates across object categories. VMP performs worse because its predefined hand trajectories offer limited flexibility. 

\noindent \textbf{Generalization Across Target Positions.}
We further assess how retrieval performance is influenced by the initial position of the target object. The target is placed in five distinct regions within the box: center, top-left, bottom-left, top-right, and bottom-right. As shown in Table~\ref{tab:real_position_generalization}, our method generalizes well across different positions, with higher success rates observed for centrally and bottom-located targets, which are closer to the robot’s base. In contrast, targets in the top corners are more challenging due to kinematic limitations and restricted manipulation flexibility.

\noindent \textbf{Real-World Retrieval Efficiency.}
We define retrieval efficiency as the time to retrieve the target object. Figure~\ref{fig:real_efficiency} compares our method against VMP and sequential grasp-and-remove. Our policy improves efficiency by leveraging skill-based actions such as pushing and stirring to expose the target object without exhaustive removal of occluders. On average, our method reduces retrieval time by 51.2\% compared to VMP and 61.9\% compared to sequential removal.

\subsection{Discussion and Limitations}\label{sec:discussion}
\noindent \textbf{Teacher-Student Gap.}
Simulation uses the privileged teacher with fingertip poses, nearby objects, and $h^{o^*}_t$ (Tables~\ref{tab:main_res}--\ref{tab:ablation_res}); deployment uses a student with only $(q_t^{arm}, x_t, y_t)$ over 15 steps. Behavior cloning transfers stirring and poking trajectories but not explicit occlusion reasoning, so spatial context about clutter remains largely implicit.

\noindent \textbf{Perception and Failures.}
SAM/Cutie tracking may drift under heavy occlusion or in top-corner settings (Table~\ref{tab:real_position_generalization}), and contact can perturb segmentation. Typical failures include tracking loss, kinematically difficult corners, and over-stirring that displaces the target.

\noindent \textbf{Efficiency, Dexterity, and Task Scope.}
Simulation efficiency (Figure~\ref{fig:retrieval_steps}) reflects the teacher's stirring strategy and dexterous hand, not the student architecture; real-world gains (Figure~\ref{fig:real_efficiency}) come from closed-loop interactions, consistent with the $r_{stir}$ ablation (Table~\ref{tab:ablation_res}). Thumb and index fingers dominate contacts, while other fingers provide compliance during stirring, unlike a parallel-jaw gripper (Table~\ref{tab:main_res}). L3 and stir ablation indicate sensitivity to clutter density; heavier occluders may still require sequential removal. 

\section{Conclusion}
We presented \emph{RetrDex}, a framework for efficient object retrieval in clutter with dexterous hands. Unlike methods that rely on sequential removal or treat pre-grasp interactions as auxiliary steps, we learn diverse interaction skills, including pushing, stirring, and poking, that expose target objects. Combining large-scale RL training in simulation with a spatially aware representation of occlusion and contact relations, RetrDex achieves efficient and robust manipulation in high-dimensional, contact-rich settings. Simulated and real-world experiments show strong retrieval performance, generalization to diverse objects, and zero-shot transfer to a physical robot. Future work includes robust perception under occlusion, distilling spatial representations to the student, and evaluation in heavier clutter.

\vspace{-0.6em}
\section*{Acknowledgements}
Partially supported by Project No.~20240313, Zhongguancun Academy, Beijing (100094). We thank anonymous reviewers for comments that improved this paper.

\bibliographystyle{IEEEtran}
\bibliography{iros2026}

\newpage
\appendices
\onecolumn

\makeatletter
\def\@IEEEprocessthesectionargument#1{%
  \@ifmtarg{#1}{%
    \@IEEEappendixsavesection*{Appendix \thesectiondis}%
    \addcontentsline{toc}{section}{\thesectiondis.}%
  }{%
    \@IEEEappendixsavesection*{Appendix \thesectiondis \\* #1}%
    \addcontentsline{toc}{section}{\thesectiondis. #1}%
  }%
}

\newcommand\appendixtableofcontents{%
  \setcounter{tocdepth}{2}%
  \begingroup
  \def\l@section##1##2{%
    \addpenalty{\@secpenalty}%
    \addvspace{0.6pc plus 1pt}%
    \begingroup
      \parindent\z@\rightskip\@pnumwidth
      \parfillskip-\@pnumwidth
      \leavevmode
      {\bfseries ##1}\titlerule*[1pc]{.}\bfseries\hbox to\@pnumwidth{\hss ##2}\par
    \endgroup
  }%
  \def\l@subsection##1##2{%
    \addvspace{0.15pc plus 1pt}%
    \begingroup
      \parindent\z@\rightskip\@pnumwidth
      \parfillskip-\@pnumwidth
      \leavevmode
      ##1\titlerule*[0.5pc]{.}\hbox to\@pnumwidth{\hss ##2}\par
    \endgroup
  }%
  \@starttoc{appendixtoc}%
  \endgroup
}

\newcommand\appendix@addtotoc[2]{%
  \addtocontents{appendixtoc}{\protect\contentsline{#1}{#2}{\thepage}{\@currentHref}}%
}

\renewcommand\addcontentsline[3]{%
  \def\appendix@tempa{#2}%
  \def\appendix@tempb{section}%
  \ifx\appendix@tempa\appendix@tempb
    \appendix@addtotoc{#2}{#3}%
  \else
    \def\appendix@tempb{subsection}%
    \ifx\appendix@tempa\appendix@tempb
      \appendix@addtotoc{#2}{#3}%
    \fi
  \fi
}
\makeatother

\setcounter{tocdepth}{2}
\setcounter{secnumdepth}{3}

\vspace{1.2em}
\begin{center}
    {\LARGE\bfseries Appendix}\\[0.6em]
    {\large\bfseries Contents}
\end{center}
\vspace{1.2em}

\appendixtableofcontents

\vspace{1em}
\newpage

\section{Implementation Details}

\subsection{Task Construction}\label{appendix:task_construction}
\noindent \textbf{Domain Randomization.} To improve robustness and generalization, we apply domain randomization during environment reset, consistent with Sec.~\ref{sec:policy_train}. At the beginning of each episode, object masses are randomized by scaling each object's default mass with a random factor sampled from $U(1, 1.5)$ (kg):
\begin{equation}
    m_\textrm{curr} = m_\textrm{default} \cdot \alpha, \quad \alpha \sim U(1, 1.5).
\end{equation}
Small perturbations are applied to the initial positions of objects (m):
\begin{equation}
    \Delta x \sim U(-0.02, 0.02), \quad \Delta y \sim U(-0.02, 0.02).
\end{equation}
The initial position of the target object within the box is also randomized (m):
\begin{equation}
    \Delta x \sim U(-0.15, 0.15), \quad \Delta y \sim U(-0.2, 0.2),
\end{equation}
which covers about 70\% of the box area. During data collection, the camera mounting position is perturbed with random displacements (m):
\begin{equation}
    \mathbf{p}_\textrm{camera} = \mathbf{p}_\textrm{default} + \boldsymbol{\epsilon}, \quad \boldsymbol{\epsilon} \sim U(-0.01, 0.01)^3.
\end{equation}

\noindent \textbf{Hand Poses during Initialization and Training.} Following Sec.~\ref{sec:task_con}, we define two static hand poses: the \emph{prepare pose}, where the hand stays above the box before retrieval, and the \emph{suspending pose}, where the hand moves away from the box. During scene initialization, the hand moves to the suspending pose while objects are dropped, then returns to the prepare pose once the clutter settles. During training, the hand may occlude the target in the top-down view; every 10 timesteps, we move it to the suspending pose to compute the pixel visibility reward before restoring the previous configuration. This periodic evaluation is disabled during policy evaluation, when rewards are not computed.

\subsection{Retrieval Policy Training}\label{appendix:rl_training}

\noindent \textbf{Reward Design.}
The reward function comprises five components: distance, stir, proximity clearance, pixel visibility, and penalty terms. Table~\ref{tab:reward_terms} lists their definitions and parameters. Following Ng et al.~\cite{10.5555/645528.657613}, we combine the first four components into a potential function (Eq.~\eqref{eq:potential}) and apply potential-based reward shaping to obtain the total reward (Eq.~\eqref{eq:reward}).

\begin{table}[H]
\centering
\renewcommand{\arraystretch}{1.2}
\setlength{\tabcolsep}{10pt}
\caption{Reward Components and Definitions.}
\label{tab:reward_terms}
\resizebox{0.85\textwidth}{!}{
    \begin{tabular}{lcc}
    \toprule
    Reward Item & Definition & Parameters \\
    \midrule
    Distance Reward & $r_\text{dist} = \exp(-5 \cdot \min(d - e_0, 0))$      & $e_0=0.15$ \\
    Stir Reward     & $r_\text{stir} = \alpha \|p^{all}_t - p^{all}_{t-1}\|_2$ & $\alpha=400.0$ \\
    Proximity Clearance Reward & $r_\text{clean} = \beta \cdot r_\textrm{dist} \cdot \sum_{i=1}^{k} d_i$ & $\beta=100.0$ \\
    Pixel Visibility Reward & $r_\textrm{pixel} = C/{15}$  & - \\
    Penalty & $\lambda_1||a^{\text{hand}}_t - a^{\text{hand}}_{t+1}||^2_2 + \lambda_2||a^{\text{arm}}_t||^2_2$ & $\lambda_1 =1.0$, $\lambda_2=1.0$\\
    \bottomrule
    \end{tabular}
}
\end{table}
Additional contact and target-displacement penalties follow Lin et al.~\cite{lin2024twisting}.

\noindent \textbf{Policy Training.} We employ the Proximal Policy Optimization (PPO) algorithm~\cite{schulman2017proximal} to train a continuous control policy using an actor-critic architecture. Detailed hyperparameters are provided in Table~\ref{tab:ppo_hyperparams}. The policy network is parameterized as a multi-layer perceptron (MLP) with three layers of sizes [256, 256, 256], utilizing the ELU activation function for improved gradient flow and non-linearity. The standard deviation of the policy distribution is learned via a log-std representation, enabling dynamic adjustment of exploration during training.

\begin{table}[H]
    \centering
    \renewcommand{\arraystretch}{1.2} 
    \setlength{\tabcolsep}{10pt} 
    \caption{Hyperparameters for PPO Training}
    \label{tab:ppo_hyperparams}
    \resizebox{\textwidth}{!}{
    \begin{tabular}{llll} 
        \toprule
        \textbf{Category} & \textbf{Parameter} & \textbf{Value} & \textbf{Description} \\
        \midrule
        \multirow{2}{*}{\makecell{\textit{Model} \\ \textit{Architecture}}} 
        & MLP Layers & [256, 256, 256] & Number of neurons per layer \\
        & Activation Function & ELU & Non-linearity used in the network \\
        \midrule
        \multirow{10}{*}{\makecell{\textit{Training} \\ \textit{Parameters}}}  
        & Learning Rate & $3\times10^{-4}$ & Step size for policy update \\
        & Discount Factor ($\gamma$) & 0.99 & Reward discounting factor \\
        & GAE Parameter ($\tau$) & 0.95 & Smoothing factor for GAE \\
        & Entropy Coefficient & 0 & Weight of entropy regularization \\
        & Gradient Clipping & Norm 1, Truncation Enabled & Prevents gradient explosion \\
        & Clip Range ($\epsilon$) & 0.1 & PPO clipping threshold \\
        & KL Threshold & 0.02 & KL divergence threshold for stopping training \\
        & Minibatch Size & 512 & Batch size for optimization \\
        & Mini Epochs & 5 & Number of updates per batch \\
        & Horizon Length & 8 & Number of steps before update \\
        & Max Training Epochs & 50,000 & Maximum number of training iterations \\
        & Value Learning Rate & $1\times10^{-3}$ & Learning rate for value function \\
        \bottomrule
    \end{tabular}
    }
\end{table}

\begin{table}[H]
    \centering
    \renewcommand{\arraystretch}{1.2}
    \setlength{\tabcolsep}{10pt}
    \caption{Hyperparameters for GNN Training.}
    \label{tab:gnn_hyperparams}
    \resizebox{\textwidth}{!}{
    \begin{tabular}{llll}
        \toprule
        \textbf{Category} & \textbf{Parameter} & \textbf{Value} & \textbf{Description} \\
        \midrule
        \multirow{4}{*}{\makecell{\textit{Model} \\ \textit{Architecture}}}
        & PointNet++ Encoder Output Dim & 64 & Dimension of point cloud encoding \\
        & GNN Hidden Dim & 64 & Hidden units in EdgeConv layers \\
        & Final Output Dim & 32 & Output dim of GNN feature extractor \\
        & Num of Neighbors (K) & 5 & Number of obstacle nodes in graph \\
        \midrule
        \multirow{4}{*}{\makecell{\textit{PointNet++} \\ \textit{Settings}}}
        & SA Layer Config & [512, 128, 1] & Points per SA layer \\
        & SA Radii & [0.2, 0.4, all] & Search radii for each SA layer \\
        & MLPs per SA Layer & [64,64,128], [128,128,256], [256,512,out] & MLPs applied in SA layers \\
        & Graph Convs & 2 EdgeConv + 1 predictor & Number and type of GNN layers \\
        \midrule
        \multirow{7}{*}{\makecell{\textit{Training} \\ \textit{Parameters}}}
        & Optimizer & Adam & Optimization algorithm \\
        & Learning Rate & $1 \times 10^{-4}$ & Initial step size \\
        & LR Scheduler & ReduceLROnPlateau & Decay LR on validation plateau \\
        & Scheduler Settings & factor 0.5, patience 5 & Scheduler configuration \\
        & Loss Function & MSE & Objective function \\
        & Batch Size & 32 & Samples per batch \\
        & Epochs & 30 & Max training epochs \\
        \midrule
        \multirow{3}{*}{\makecell{\textit{Validation} \\ \textit{Settings}}}
        & Validation Interval & Every 29 steps & Frequency of validation \\
        & Data Split & 90\% train / 10\% val & Dataset partition ratio \\
        & Time Sequence Mode & \texttt{last} (default) & Sampling strategy from sequence \\
        \bottomrule
    \end{tabular}
    }
\end{table}

To ensure stable learning, we adopt adaptive learning rate scheduling, starting at $3\times10^{-4}$. The advantage function is normalized to reduce variance in policy gradient updates. The Generalized Advantage Estimation (GAE)~\cite{SchulmanMLJA15} parameter $\tau$ is set to 0.95. Gradient clipping with a norm threshold of 1 is applied to prevent exploding gradients. We use a PPO clipping range of 0.1 and a KL divergence threshold of 0.02 to limit large policy updates.

The training runs for a maximum of 50,000 epochs, with model checkpoints saved every 1,000 epochs. The best-performing model is selected based on validation returns and retained after 200 epochs to prevent overfitting. A separate centralized value function is used for advantage estimation, parameterized as an MLP with the same architecture as the policy network. The critic network employs a higher learning rate of $1\times10^{-3}$ to facilitate faster convergence in value estimation, a choice informed by preliminary experiments indicating more stable critic updates with this configuration.

\subsection{Spatially Aware Representation Training}\label{appendix:gnn_training}
\noindent \textbf{Model Architecture.}
We design a graph neural network (GNN) module to predict target visibility and extract a spatially aware representation, following Sec.~\ref{sec:policy_train}. Each node is represented by a sampled point cloud $\mathcal{P}_i \in \mathbb{R}^{256 \times 3}$ and encoded into a 64-dimensional feature vector using PointNet++~\cite{DBLP:conf/nips/QiYSG17}. We build a complete graph over the target node, the hand node, and $K=5$ nearest neighboring objects, yielding $K+2$ nodes in total. Two EdgeConv~\cite{WangSLSBS19} layers propagate features among nodes, and a third EdgeConv layer aggregates pairwise edge features for visibility prediction. The target node embedding after message passing, denoted $h^{o^*}_t$, serves as the spatially aware representation injected into the policy observation space.

\noindent \textbf{Implementation Details.}
Before training this module, we collect 2{,}000 trajectories from a basic RL policy. We optimize the network with mean squared error (MSE) to regress predicted visibility against the number of target pixels visible in the top-view segmentation mask. The dataset is split into 90\% for training and 10\% for validation. We use the Adam optimizer with a learning rate scheduler that reduces the learning rate when validation loss plateaus. Validation is performed every 29 training iterations.

Each episode contains 210 timesteps, and the model is trained on either the last step or all timesteps, depending on the configuration. Table~\ref{tab:gnn_hyperparams} summarizes the hyperparameters. The output dimensionality of the extracted feature used in downstream RL is set to $d=32$ unless otherwise specified.

\subsection{Sim-to-Real Transfer}\label{appendix:sim_real_transfer}
For real-world deployment, we collect rollout trajectories from the trained teacher policy and distill a student policy via behavior cloning (BC)~\cite{NIPS1988_812b4ba2}, following Sec.~\ref{sec:sim2real}. We keep successful trajectories in which the fingers maintain at least 2 cm clearance above the box bottom, and we balance target positions within the box to avoid overfitting to specific placements.

\begin{wrapfigure}{r}{0.4\linewidth}
    \vspace{-.9em}
    \centering
    \includegraphics[width=0.99\linewidth]{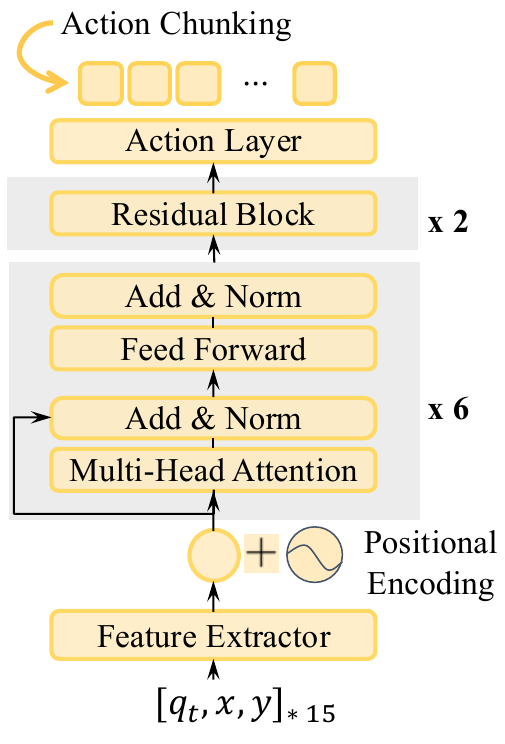}
    \caption{Architecture of the distilled student policy trained via behavior cloning.}
    \label{fig:student_pi}
\end{wrapfigure}

Figure~\ref{fig:student_pi} shows the student architecture. The observation at each timestep is $o_t = (q_t^{arm}, x_t, y_t)$, consisting of arm joint positions and the target object's pixel coordinates in the top-down camera frame. The policy processes a history of 15 timesteps and outputs actions $a_t \in \mathbb{R}^{13}$ for the arm and dexterous hand. In the real world, the target is tracked online with SAM~\cite{kirillov2023segment} and Cutie~\cite{cheng2024putting}, and pixel coordinates are converted to world coordinates via camera calibration at 30 Hz. The network uses six transformer layers with six attention heads, a hidden dimension of 384, and a feed-forward dimension of 2048. We train the student with the Adam optimizer at a learning rate of $1 \times 10^{-4}$ for 10{,}000 iterations and a batch size of 512. Mixed-precision training and gradient clipping at 1.0 are used for stable optimization. Table~\ref{tab:student_policy_parameters} lists the architectural and training hyperparameters. 

\subsection{Baseline Implementation}\label{appendix:baseline}
In our simulation experiments, we compare our method against several baseline approaches defined in the main paper. \textit{Ours (SAC)} replaces PPO with Soft Actor-Critic (SAC)~\cite{haarnoja2018soft} while keeping all other components identical. \textit{Ours (Gripper)} substitutes the dexterous hand with a parallel-jaw gripper to assess the impact of end-effector morphology. The other two baselines are \textit{Visual-based Motion Planning (VMP)} and \textit{Grasp-Pick}. VMP uses target segmentation masks to guide a scripted motion planner with hand-designed rules. Grasp-Pick sequentially grasps and removes occluding objects based on support relationships using ground-truth object poses.

\subsubsection{Ours (SAC)}
SAC is an off-policy reinforcement learning algorithm grounded in the maximum entropy framework, which promotes exploration by encouraging stochastic action selection. The algorithm comprises a soft Q-function $Q_\theta(s,a)$ and a stochastic policy $\pi_\phi(a|s)$.

The soft Q-function is defined as the expected cumulative return:
\begin{equation}
  Q_\theta(s_t, a_t) = \mathbb{E}\left[\left. \sum_{t'=t}^{T} \gamma^{t'-t} r_{t'} \right| s_t, a_t \right],
\end{equation}
where $\gamma \in [0,1]$ denotes the discount factor. In maximum entropy RL, this return is typically augmented with an entropy bonus, though omitted here for clarity.

The Q-function parameters $\theta$ are optimized by minimizing the soft Bellman residual:
\begin{align}
  J_Q(\theta) = \mathbb{E}_{(s_t, a_t, s_{t+1}, r_t) \sim \mathcal{B}} \left[\left(Q_\theta(s_t, a_t) - r_t - \gamma {\bar V}(s_{t+1}) \right)^2 \right],
\end{align}
where ${\bar V}(s_{t+1}) = \mathbb{E}_{a \sim \pi_\phi} \left[ Q_{\bar{\theta}}(s_{t+1}, a) - \alpha \log \pi_\phi(a|s_{t+1}) \right]$ uses a target Q-network with parameters $\bar{\theta}$. The temperature $\alpha$ controls the strength of the entropy regularization and can be fixed or learned during training.

The policy parameters $\phi$ are updated by minimizing:
\begin{equation}
  J_\pi(\phi) = \mathbb{E}_{s_t \sim \mathcal{B}, a_t \sim \pi_\phi} \left[ \alpha \log \pi_\phi(a_t|s_t) - Q_\theta(s_t, a_t) \right].
\end{equation}

SAC alternates between policy evaluation and improvement. In the \textit{Ours (SAC)} baseline, we replace PPO with SAC while keeping the observation space, reward design, and training pipeline unchanged.

\begin{table}[!t]
    \centering
    \renewcommand{\arraystretch}{1.2} 
    \setlength{\tabcolsep}{10pt} 
    \caption{Hyperparameters of Distilled Policy.}
    \label{tab:student_policy_parameters}
    \resizebox{0.9\linewidth}{!}{
    \begin{tabular}{llll}
    \toprule
    \textbf{Category} & \textbf{Parameter} & \textbf{Value} & \textbf{Description} \\
    \midrule
    \multirow{9}{*}{\makecell{\textit{Model} \\ \textit{Architecture}}} 
     & Input State Dimension & 9 & Size of input state vector \\
     & Action Dimension & 13 & Number of output actions \\
     & History Frames & 15 & Past frames used as input \\
     & Future Action Frames & 5 & Future actions predicted \\
     & Transformer Hidden Size ($d_{model}$) & 384 & Hidden layer size \\
     & Number of Attention Heads & 6 & Transformer attention heads \\
     & Number of Transformer Layers & 6 & Transformer depth \\
     & Feed-forward Dimension & 2048 & FFN hidden size \\
     & Dropout Rate & 0.15 & Dropout probability \\
    \midrule
    \multirow{6}{*}{\makecell{\textit{Training} \\ \textit{Parameters}}}  
     & Batch Size & 512 & Training batch size \\
     & Total Iterations & 10,000 & Training iterations \\
     & Learning Rate & 1e-4 & Initial learning rate \\
     & Optimizer & Adam & Optimization algorithm \\
     & Loss Function & Behavior cloning (BC) & Objective for distillation \\
     & Gradient Clip Norm & 1.0 & Gradient clipping threshold \\
    \bottomrule
    \end{tabular}
    }
\end{table}

\begin{table}[!t]
    \centering
    \renewcommand{\arraystretch}{1.2} 
    \setlength{\tabcolsep}{10pt} 
    \caption{Hyperparameters for SAC Training}
    \label{tab:sac_hyperparameters}
    \resizebox{0.9\linewidth}{!}{
    \begin{tabular}{llll}
    \toprule
    \textbf{Category} & \textbf{Parameter} & \textbf{Value} & \textbf{Description} \\
    \midrule
    \multirow{4}{*}{\textit{Network Architecture}} & Actor Hidden Dim & 256 & Hidden layer dimension of actor network \\
    & Actor Hidden Depth & 3 & Number of hidden layers in actor network \\
    & Critic Hidden Dim & 256 & Hidden layer dimension of critic network \\
    & Critic Hidden Depth & 3 & Number of hidden layers in critic network \\
    \midrule
    \multirow{15}{*}{\textit{Training}} & Actor Learning Rate & 5e-4 & Learning rate for actor network \\
    & Critic Learning Rate & 5e-4 & Learning rate for critic network \\
    & Temperature Learning Rate & 0.05 & Learning rate for entropy coefficient \\
    & Batch Size & 8192 & Number of samples per training batch \\
    & Max Gradient Norm & 0.5 & Maximum gradient norm for clipping \\
    & SAC Epochs & 8 & Number of training epochs per update \\
    & Discount Factor & 0.99 & Future reward discount factor \\
    & Initial Temperature & 0.1 & Initial entropy coefficient \\
    & Critic Tau & 0.05 & Soft update coefficient for target critic \\
    & Learnable Temperature & True & Whether to learn entropy coefficient \\
    & Number of Environments & 512 & Parallel environments in IsaacGym \\
    & Update Interval & 1 & Environment steps between updates \\
    & Seed Steps & 32 & Initial random steps per environment \\
    & TD n-step & 3 & Number of steps for TD learning \\
    & Replay Buffer Size & 5M & Maximum capacity of experience replay \\
    \bottomrule
    \end{tabular}}
\end{table}

We implement the baseline \textit{Ours (SAC)}, incorporating several key modifications to improve training efficiency and stability. The implementation is built on the IsaacGym physics engine, enabling large-scale parallelism with 512 simulated environments to accelerate data collection. The full set of hyperparameters is listed in Table~\ref{tab:sac_hyperparameters}. Both the actor and critic are modeled as three-layer multilayer perceptrons, each with 256 hidden units per layer. The actor outputs the mean and log standard deviation of a Gaussian action distribution, with the log standard deviation constrained between -5 and 2 to ensure numerical stability. The critic adopts a double Q-learning architecture with target networks updated via a soft update rule using $\tau = 0.05$.

To stabilize training, we use a large batch size of 8192 and apply policy updates at every timestep. At the start of training, each environment performs 32 random actions to populate the replay buffer. The actor and critic are optimized using the AdamW optimizer with a learning rate of 5e-4, while the temperature parameter is updated with a learning rate of 0.05. Gradient clipping with a threshold of 0.5 is applied to mitigate exploding gradients. The replay buffer stores up to 5 million transitions. For each policy update, we perform 8 epochs of optimization using a single mini-batch per epoch, promoting efficient data usage while reducing the risk of overfitting. The temperature parameter is initialized to 0.1 and is learned adaptively to maintain a balance between exploration and exploitation.

Our parallelized training framework significantly enhances sample throughput by synchronously simulating 512 environments, all of which share a single policy network. This design ensures consistent policy updates while leveraging high-throughput simulation. Additionally, we adopt 3-step temporal difference (TD) learning to reduce bias in value estimation and improve overall policy performance.

\subsubsection{VMP} The VMP system implements a vision-guided manipulation framework for dexterous robotic retrieval tasks in cluttered environments. It integrates visual perception, motion planning, and control execution through a state machine architecture to ensure reliable object manipulation.

The vision module employs a top-down camera with a resolution of $1024 \times 512$, capturing RGB, depth, and segmentation maps of the workspace. Target objects are identified using segmentation masks obtained from the segmentation map, with their IDs corresponding to known object labels. The center of the target mask’s bounding box is extracted as the 2D image coordinate, which is projected into 3D space using depth data to obtain precise object localization. For motion planning, the robotic arm moves its end effector to the computed 3D coordinate and performs a scrape action to retrieve the object. 

When the target object is completely occluded and its segmentation mask cannot be detected, the system employs an exploration strategy by randomly sampling four 3D coordinates within the cluttered bin area. The arm sequentially moves to these coordinates, performing scrape actions to uncover the target object. Specifically, the entire motion planning and scrape action process employs a four-stage approach to ensure reliable object retrieval.

\textbf{Pre-approach stage:} The end-effector moves to a predefined position ($h = 0.5\,\textrm{m}$) above the target object. This configuration facilitates subsequent control of the hand to reach any position within the bounding box.

\textbf{Final approach stage:} Precise positioning is achieved using visual feedback combined with damped least squares inverse kinematics:
\begin{equation}
    \tau = J^T (JJ^T + \lambda I)^{-1} \Delta x \notag
\end{equation}
where $\lambda = 0.05$ is the damping parameter, $J$ is the Jacobian matrix, and $\Delta x$ represents the positional error.
    
\textbf{Scraping stage:} The system executes a periodic motion pattern defined by:
\begin{equation}
    x(t) = A \sin(2\pi f t + \phi) + O \notag
\end{equation}
where the amplitude $A = 2.0$, frequency $f = 20\,\textrm{Hz}$, phase shift $\phi = \pi/4$, and offset $O = 0.5$.

\textbf{Reset stage:} The target object has been retrieved, so the robot arm will return the end effector to its initial position.

The control execution module utilizes position-based control for both arm and finger joints. Adaptive damping parameters are applied to ensure stable motion, while joint limits are strictly enforced throughout the execution process: $q_{\textrm{min}} \leq q \leq q_{\textrm{max}}$.

\subsubsection{Grasp-Pick} This method relies on support relationships among cluttered objects to guide the grasping sequence. To ensure these assumptions hold, we designed tailored setups for both simulation and real-world experiments.

In simulation, object positions are directly accessible, enabling precise calculation of support relationships. We employ a KD-Tree~\cite{skrodzki2019kd} to organize the coordinates of objects near the target. Based on Euclidean distance, we select the three to five nearest objects, depending on the scenario, and manipulate them sequentially to clear access to the target. This approach is computationally simple but relies on accurate object poses and may not generalize to complex clutter.

For robotic control, we implement damped least squares inverse kinematics: 
\begin{equation} 
    \dot{\mathbf{q}} = \mathbf{J}^T(\mathbf{J}\mathbf{J}^T + \lambda \mathbf{I})^{-1}\dot{\mathbf{x}}, \notag 
\end{equation} 
where $\lambda$ is the damping coefficient, $\mathbf{J}$ is the Jacobian matrix, and $\dot{\mathbf{x}}$ is the desired end-effector velocity. This formulation offers stable solutions near singularities but may limit the dexterity needed in cluttered environments.

In real-world experiments, sequentially grasping and removing multiple objects in stacked scenes with a dexterous hand remains challenging due to perception and control limitations. To address this, we employ predefined trajectories for each trial, simulating an idealized execution scenario. While this implementation provides an upper-bound estimate of this method’s efficiency, it does not reflect the challenges of autonomous execution in unstructured environments.

\subsection{Evaluation Metrics}\label{appendix:eval_metric}
\noindent \textbf{Exposure Calculation.}
We evaluate retrieval effectiveness and efficiency using pixel-level target visibility, as defined in the Experiments section of the main paper. \textit{Exposure} is defined as the ratio between the number of visible target pixels under occlusion ($p_t^\textrm{curr}$) and the fully visible pixel count ($p_t^\textrm{all}$) at the same 6D pose. Specifically, we record the target pose and visible pixel count every 10 steps during an episode. After the episode, all occluding objects are removed in simulation, the target is rendered at the recorded pose, and $p_t^\textrm{all}$ is computed. The exposure at timestep $t$ is:
\begin{equation}
    \textrm{exposure}_t = \frac{p_t^\textrm{curr}}{p_t^\textrm{all}}.
\end{equation}

\begin{figure}[!t]
    \centering
    \includegraphics[width=0.95\linewidth]{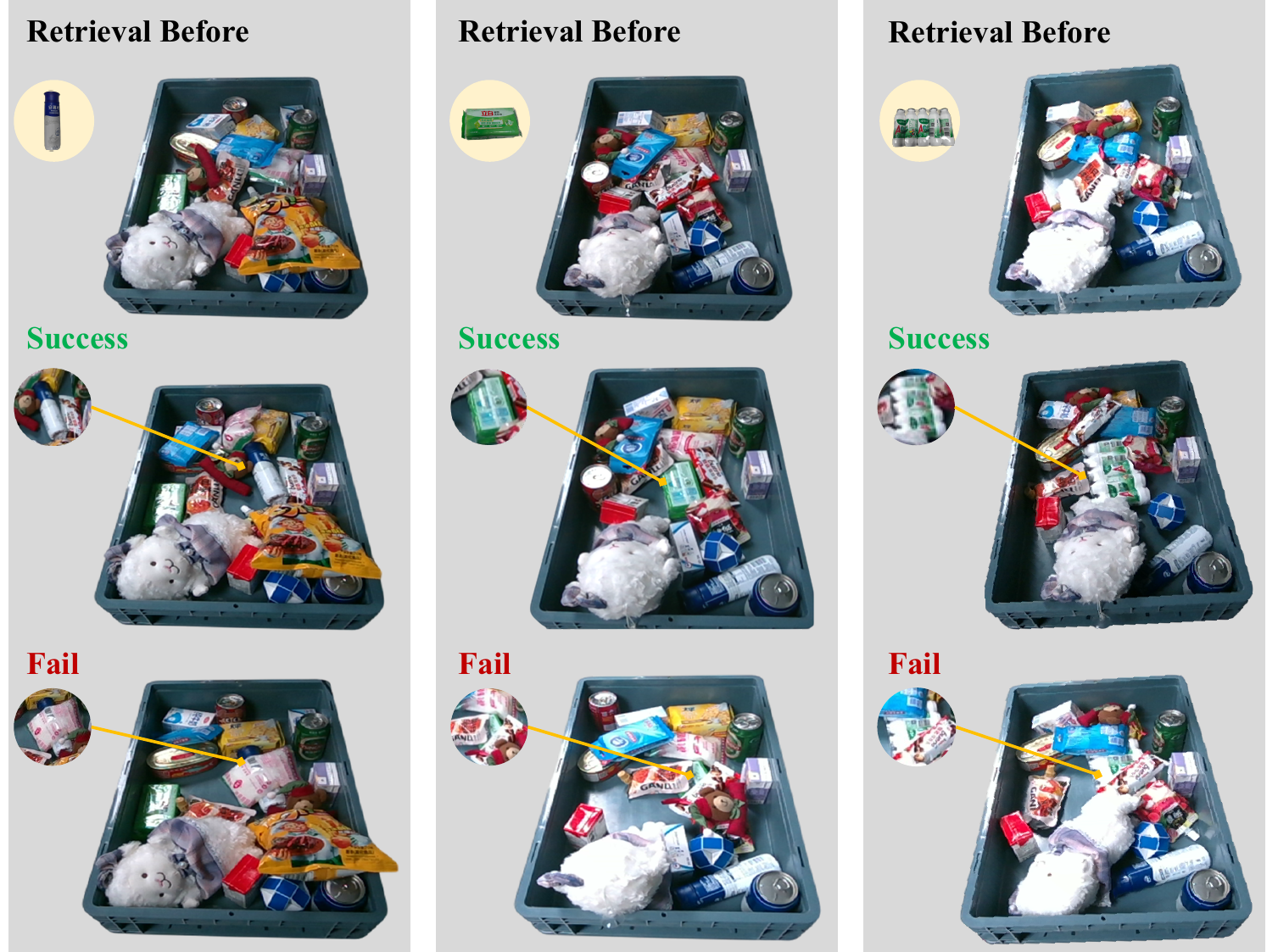}
    \caption{\textbf{Examples of successful and failed object retrievals on the real robot.}}
    \label{fig:real_eval_metric}
\end{figure}

\noindent \textbf{Success Criteria and Real-World Evaluation.}
A retrieval is successful if the final exposure exceeds 95\% within 210 steps. We report retrieval success rate (RSR), retrieval steps (RS), and increase in exposure ratio (IER), as defined in the main paper. On the real robot, we capture images before and after each trial with a side-mounted RealSense D435 camera and compare target exposure. Figure~\ref{fig:real_eval_metric} shows examples of successful and failed retrievals.

\subsection{Computing Resources}
All experiments are conducted on a single NVIDIA RTX 4090 GPU. The reinforcement learning training takes approximately 6 hours, the GNN module requires around 20 minutes, and the student policy training completes within 4 minutes.

\section{Additional Experiments}
\subsection{Impact of Occlusion Rate}
\begin{figure}[H]
    \centering
    \includegraphics[width=0.5\linewidth]{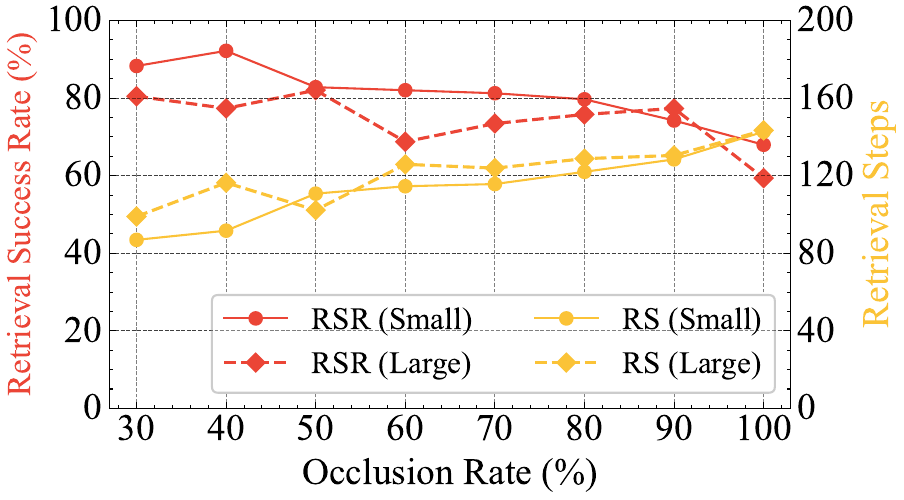}
    \caption{\textbf{Impact of Occlusion Rate on Performance and Efficiency.} We evaluate the retrieval success rate and retrieval steps of our policy for small and large target objects under varying occlusion levels.}
    \label{fig:occlusion}
\end{figure}

We study how clutter occlusion rate ($1-\textrm{exposure}$) affects retrieval performance. Figure~\ref{fig:occlusion} reports RSR and RS for small and large target objects under varying occlusion levels. As occlusion increases from 30\% to 100\%, RSR decreases for both object sizes, and RS increases, indicating lower success and efficiency in heavier clutter. Smaller objects achieve higher RSR and fewer steps than larger ones at most occlusion levels, likely because larger targets require clearing more surrounding clutter.

\end{document}